\documentclass{mypaper}

\newcommand{\phn}[1]{\textsc{#1}}
\newcommand{\wrd}[1]{\textit{#1}}

\usepackage{booktabs}
\usepackage{tabularx}
\newcommand{\mytable}{
    \footnotesize
    \centering
    \renewcommand{\arraystretch}{1.2}
}
\newcolumntype{C}{>{\centering\arraybackslash}X}
\newcolumntype{L}{>{\raggedright\arraybackslash}X}
\newcolumntype{R}{>{\raggedleft\arraybackslash}X}
\newcolumntype{P}[1]{>{\raggedright\arraybackslash}p{#1}}
\usepackage{etoolbox}
\newcommand{\ubold}{\fontseries{b}\selectfont}
\robustify\ubold

\title{On the information rate of phones, syllables and words}
\author{Herman Kamper}
\address{Electrical and Electronic Engineering, Stellenbosch University, South Africa}

\begin{document}

\maketitle

\section*{Abstract}

Which linguistic units minimise the bitrate required to encode speech content: phones, syllables, words, maybe larger units, or something in between? This question has implications for the development of speech and language systems and theories of human language processing. Using a transcribed corpus of 960 hours of English audiobooks, I compare the information rates of phones, syllables, words, multi-word combinations and intermediate tokenisations. While all units are equivalent under a perfect language model, it makes sense to compare these under limited modelling budgets; I use $n$-gram and transformer language models. The lowest bitrates are achieved using words, with slightly longer units sometimes performing best. The results show that, when modelling capacity is limited, units larger than syllables but smaller than long multi-word phrases provide the most efficient representation in terms of information rate.

\section{Introduction}

\begin{figure}[b]
    \centering
    \includegraphics{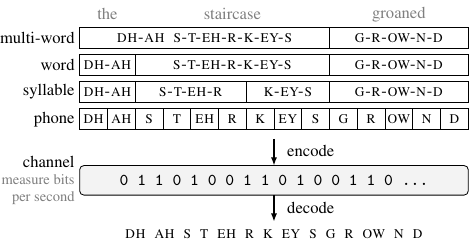}
    \caption{This work compares the information rates of different linguistic units when used to encode the same phonetic content.}
    \label{fig:setup}
\end{figure}

Suppose I have a spoken utterance and I want to send you the content of that utterance. % (the speech sounds it contains).
I write the utterance down phonetically, I encode the phones in bits, you receive the bits, and (because we agreed beforehand on an encoding scheme) you can decode the bits and know what speech sounds were in the utterance. For example, if the utterance is \wrd{the staircase groaned}, I would encode \phn{dh ah s t eh r k ey s g r ow n d}, as shown in the phone tier of Figure~\ref{fig:setup}. For our encoding, we assign bit sequences to each phone unit based on the probability of the unit, meaning our scheme uses the minimum number of bits on average~\cite{shannon1948mathematical,cover2006information,sanderson2026entropy}. We measure how many bits this optimal phone-based encoding~uses.

What if, rather than phones, we decide to tokenise the utterance into syllables? The goal is still to transmit the phonetic content, but we just do it with a different set of units, e.g.\ \phn{dh-ah\ \  s-t-eh-r\ \ k-ey-s\ \  g-r-ow-n-d}. Using the optimal syllable-to-bits encoding scheme, I again encode the units, send the bits, you decode the bits into syllables, and then you decode the syllables into phones. A different tokenisation, but we've communicated exactly the same content. Is this more efficient than the phone encoding? Rather than syllables, why not words? %Now we will have a single unit \phn{s-t-eh-r-k-ey-s} (\wrd{staircase}).
Or maybe we can communicate even more efficiently with multi-word units like \phn{dh-ah-s-t-eh-r-k-ey-s} (\wrd{the staircase}); if in my data these two words often co-occur, I could save bits by encoding these together (Figure~\ref{fig:setup} top). By now you probably get the~picture.

This paper asks which units -- phones, syllables, words, units longer than words, something in between -- are the most efficient for communicating linguistic content. Experiments are done on transcribed phonetic sequences (from forced alignments) on English data (audiobooks from the LibriSpeech~\cite{panayotov2015librispeech} corpus).

Why do this comparison? First, the literature does not report the relative efficiencies of these different units. Studies report bitrates like 50 bits per second (bps) for phones~\cite{lopes2010alternatives,flanagan2014speech} or 39 bps for syllables~\cite{coupe2019languages}. But these are not directly comparable since information rate is defined differently in each study and is not estimated on the same data.%\footnote{In defence of Coupé et al.~\cite{coupe2019languages}, their goal was not to produce \textit{the} bitrate of syllables, but rather to compare information rates across different languages when encoding the same (translated) content.}
%To make a direct comparison, information rates have to be measured consistently and on the same data.

A second motivation for such a comparison is its relevance
to human language processing, which is likely shaped by pressures for efficient communication~\cite{gibson2019efficiency}. I do not want to state this too strongly, because this paper presents a purely information-theoretic analysis ignoring human mental effort, how the mental message is converted into a phonetic sequence, production and perception effort, and channel noise. Nevertheless, characterising relative efficiency is relevant to 
%characterising the relative information rates of different linguistic units may help inform
theories of linguistic representation, such as how the mental lexicon is organised~\cite{virpioja2018statistical}.

The third motivation is for the zero-resource speech processing community. Over more than a decade, this community has tried to learn linguistically meaningful units from unlabelled speech without any transcriptions~\cite{dunbar2021zero}. Studies have looked at the discovery of phone-like units~\cite{lee2015unsupervised,cuervo2022contrastive}, syllabic units~\cite{rasanen2018prelinguistic,cho2025sylber,baade2025syllablelm}, words~\cite{park2008unsupervised,malan2026should}, and even phrase-like units~\cite{vanniekerk2024spokenterm}. But why should we be interested in particular units rather than others? None of this paper deals with speech (all experiments are on transcribed phonetic sequences). But still, as guidance for this community, it is worth knowing whether some units are more efficient than others in an idealised communication setting.

After defining information rate, I first show that all units, in fact, have the same bitrate under a perfect model: different tokenisations of the same input are just different encodings of the same underlying events.
But this is only true with a perfect model -- something we never have. I therefore refine the research question to consider the different information rates when using real models to estimate the probabilities. The models are $n$-gram language models~\cite{chen1999empirical} and transformer language~models~\cite{vaswani2017attention}.

Apart from phones, syllables and words, I also apply byte pair encoding (BPE)~\cite{gage1994new} at the word-token level, generating multi-word units, like the one for \textit{the staircase} in the top tier of Figure~\ref{fig:setup}.
Larger units are always expressed in terms of their phones to ensure a one-to-one correspondence between tokenisations, as shown in the figure.
Intermediate sanity checks, like applying subword BPE on phone units, are also done. 

I find that the best bitrates are achieved with a 20M-parameter transformer using word units (28.07 bps). With $n$-grams, adding a small number of word-BPE merges improves results over isolated words. % (30.97 vs 31.05 bps).
For both model classes, if too many word merges are added, bitrate gets worse. Under a constrained modelling budget, it therefore seems that word-like units strike a good balance, sitting between a small set of shorter-duration units (e.g.\ phones; best bitrate 29.56 bps) vs having a larger set of longer units (e.g.\ word-BPE combinations; 29.27 bps when adding 5,000 word merges).
Code is released at \url{https://github.com/kamperh/inforate/}.

\section{Information rate}

Information rate, measured in bits per second, is defined as~\cite{coupe2019languages}:
\begin{equation*}
    \text{bitrate}
    = \frac{\text{bits}}{\text{duration}}        
    = \frac{\text{\#units}}{\text{duration}} \times \text{bits per unit}
\end{equation*}
To know the average bits per unit, we need to decide on an encoding scheme to assign bit sequences to unit types. If we use entropy coding, where the number of bits is chosen based on the probability of the unit, then the bits per unit would be given by the entropy of the distribution over the units~\cite{shannon1948mathematical,cover2006information,sanderson2026entropy}. But, to do this, we would need to know the true unit probabilities $P(w_t | w_{1:t-1})$.
We don't have these. Instead we have a model $P_{\theta}(w_t | w_{1:t-1})$.\footnote{I use $w$ here, but these units don't need to be words. The whole point is that we will look at different units.} Under this model, we can then estimate the number of bits on held-out data (consisting of $T$ tokens):
\begin{equation*}
    \text{bits per unit} = - \frac{1}{T} \sum_{t = 1}^T \log_2 P_{\theta}(w_t | w_{1:t-1})
\end{equation*}
Formally, this is the estimated cross-entropy rate when we use model $P_{\theta}$ to encode units coming from the true distribution $P$. We can now measure the bitrate under the model:
\begin{equation}
    \label{eq:bitrate}
    \text{bitrate}
    = \frac{\text{bits}}{\text{duration}}
    = -\frac{1}{D} \sum_{t = 1}^T \log_2 P_{\theta}(w_t | w_{1:t-1})
%    = -\frac{1}{D} \log_2 P_{\theta}(w_{1:T})
\end{equation}
where $D$ is the data duration in seconds.

\vspace{5pt}
\noindent \textbf{Research question.} What are the relative coding efficiencies of different linguistic units (phones, syllables, words, etc.)? Essentially we are asking what $w$ should be to give low bitrates.

\vspace{5pt}
\noindent 
\textbf{Other bitrate calculations.} Bitrate has been calculated differently elsewhere. The Zero Resource Speech Challenges, for instance, use~\cite{dunbar2019zero}:
\begin{equation*}
    \text{bitrate} = - \frac{ T }{D} \sum_{w} P_{\theta}(w) \log_2 P_{\theta}(w)
\end{equation*}
Here $P_{\theta}(w)$ are unigram probabilities estimated from counts on the evaluation data; bitrate is taken as the expected value under this distribution. Coupé et al.~\cite{coupe2019languages} do something similar, but with a second-order model that takes the preceding token into account.
%:
%\begin{equation*}
%    \text{bitrate}
%    = -\frac{T}{D} \sum_{w_{t - 1}} \sum_{w_{t}} P_{\theta}(w_{t - 1}, w_t)
%    \log_2 \frac{ P_{\theta}(w_{t - 1}, w_t) }{ P_{\theta}(w_{t - 1})}
%\end{equation*}
Both these match~\eqref{eq:bitrate} if we (a) make a first-/second-order $n$-gram modelling assumption, and (b) calculate bitrate on the same data used for the unigram/bigram model estimates. This is reasonable for coarse comparisons, but (a) units can presumably be encoded much more efficiently with larger contexts using a higher capacity $P_{\theta}(w)$, and (b) these measurements do not account for generalisation to unseen data.

\section{But units don't matter: they all have the same ideal bitrate}
\label{sec:units_dont_matter}

We now have an appropriate way to measure information rate. But there is still a question of what the model $P_{\theta}$ should look like. I now briefly digress to show that, under an ideal model, all units will have the same bitrate. But a comparison is still meaningful if the model $P_{\theta}$ is constrained.

%The experiments in this paper consistently use phones as the smallest atomic unit from which larger units are built. But f
For illustration, let us assume we are using orthographic words as units, and are wondering whether we should not rather use larger units. For instance, maybe in our data the word \wrd{York} is always preceded by \wrd{New}. Can't we save bits by having one unit \wrd{New York} instead of two separate units \wrd{New} and \wrd{York}, each separately taking up bits? No, it doesn't matter how you tokenise the input; the following should hold:
\begin{equation*}
\begin{aligned}
    &P(\textit{New York is beautiful}) \\
    &\quad= P(\textit{New}) \cdot P( \textit{York} | \textit{New}) \cdot P( \textit{is} | \textit{New York}) \cdots \\
    &\quad= P(\textit{New York}) \cdot P( \textit{is} | \textit{New York}) \cdots
\end{aligned}
\end{equation*}
But how does this make sense, given that we should have saved bits? % by merging \wrd{New} and \wrd{York}?
Turns out that if \wrd{York} always followed \wrd{New} (and no other word did), then $P(\textit{York}|\textit{New}) = 1$. And so $P(\textit{New York}) = P(\textit{New})$. In information-theoretic terms: it takes zero extra bits to transmit \wrd{York}. More formally, different tokenisations correspond to different random variables defined on the same underlying sample space.

So, apparently, the question of the relative information rates of different units is silly -- they are all the same. But this relies on a strong assumption: that we know the true probability distribution $P$, the actual distribution governing the world. But no one (except the Creator) actually knows these probabilities. So we always have a model $P_{\theta}$. This model will be finite. And this model would be tied to a particular type of unit. And now, under this limited capacity, it matters how you tokenise your input.

\vspace{5pt}
\noindent \textbf{Research question refined.} What are the relative coding efficiencies of different linguistic units \textit{under a limited modelling budget}?

\section{Models}

I consider two classes of finite models: $n$-gram language models~\cite{chen1999empirical,jurafsky2026speech} and transformer-based neural language models~\cite{vaswani2017attention}.

\subsection{Classic $n$-gram language models}

For the $n$-gram models, I use absolute discounting in which higher-order $n$-gram language models are interpolated with lower-order models~\cite{chen1999empirical}. Discounted mass is used to deal with unit types that are unseen at training time.

Formally, for the highest order $n$, we have~\cite[\S2.6]{chen1999empirical}:
\begin{equation*}
    P_n(w_t | h) = \frac{ \max \{ c(h, w_t) - d, 0\} } {c(h)} +  \frac{d N_{1+}(h\cdot)} { c(h) } P_{n - 1}(w_t | h')
\end{equation*}
where $d$ is the discount value, $h = w_{t - n + 1 : t - 1}$ denotes the preceding units, $h' = w_{t - n + 2 : t - 1}$ drops the first symbol from $h$, and $N_{1+}(h\cdot) = |\{w : c(h, w) > 0\}|$ is the number of distinct continuations after $h$. This recurses down to the unigram case:
\begin{equation*}
P_1(w_t) = \frac{ \max \{ c(w_t) - d, 0\} } {T} + \frac{Vd}{T} P_0 \ , \quad P_0 = \frac{1}{V + 1}
\end{equation*}
The $+1$ in the denominator of $P_0$ reserves probability mass for unseen types.

Instead of a single discount value $d$ for all orders, these are set per order~\cite{ney1994structuring}: $d = n_1 / (n_1 + 2n_2)$, with $n_1$ and $n_2$ the number of distinct $n$-grams occurring exactly once and exactly twice in the training data (at this order). If these are zero, $d = \text{0.75}$ is used as a fall-back (I did not tune this value). 
%I also clamp t
The discount is also clamped to the range $d \in [\text{0.1}, \text{0.9}]$ to avoid strange~values.

When dealing with syllables and words, we have to deal with units that are not seen during training.
For these, I back off to a spelling model, paying bits to spell out the unit token into its constituent phones. So if we have a token \phn{g-r-ow-n-d} (\wrd{groaned}) that has never been seen, it will be decomposed into five tokens \phn{g r ow n d} (with an end-of-word symbol). This spelling model is itself an absolute discounted $n$-gram using the interpolation as above, but trained on spelled-out tokens. 

\vspace{5pt}
\noindent 
\textbf{Implementation.}
In development experiments, I varied the $n$-gram order (from 1 to 5) for all the different units. For most of the units, trigrams worked the best, and I therefore use this throughout. I also always use a third-order spelling model. This is then our fixed $n$-gram modelling budget.

\subsection{Transformer language models}

I use a decoder-only autoregressive transformer language model based on a modification of the OPT-125M architecture~\cite{zhang2022opt}. I reduced the model size so that the body (excluding the output unit-predicting layer) has approximately 20M parameters. This is kept constant regardless of the input unit.\footnote{I did not tune model size. The number of tokens per parameter differs based on the unit, but for phones, this gives 1.69 tokens per parameter. Later you see that I do also check bitrates with a much bigger separately trained Qwen model~\cite{qwen2024qwen2.5}. This sanity check gives confidence that the budget here is reasonable.} Each model is a standard causal transformer using stacked blocks with masked multi-head self-attention and a feedforward layer, pre-layer-norm, ReLU activations, learned positional embeddings, and tied input and output embeddings. I use 6 transformer layers with 512 hidden dimensions, 8 heads, feedforward networks with 2048 dimensions, and a context window of 2048 tokens. The output layer is different depending on the input unit. More detail is given below, but the phone neural language model only has 39 outputs (together with sentence-boundary and pad tokens) while the word model has roughly 56k outputs.\footnote{You could argue that I should have balanced model sizes taking the output layer into account. But it is not clear exactly how this should be done. Also note that the word and all the word-BPE models have roughly the same number of outputs (details to follow).}

\vspace{5pt}
\noindent 
\textbf{Implementation.}
The neural models are trained with an effective batch size of 40 on a single NVidia RTX3070 (8~GB). I use AdamW optimisation~\cite{loshchilov2019decoupled} with a linear warmup for 8\% of the steps to a maximum learning rate of $\text{1}\cdot\text{10}^{-3}$ followed by a cosine decay to $\text{2}\cdot\text{10}^{-5}$. % over the remaining steps. 
Models are trained for approximately 20 epochs; the step sizes are changed based on the unit to ensure that the equivalent of roughly 20 passes is made over the encoded data. All these values are based on the recipe from~\cite{visser2026zerosyl}. %The only change was to raise the maximum learning rate from $\text{5}\cdot\text{10}^{-4}$ to $\text{1}\cdot\text{10}^{-3}$; this slightly improved final performance across all the units (but the trends were always the same). I also experimented with training for more epochs, but this didn't help any of the units. No further development experiments were done.

\section{Data and units}

I perform experiments on the multi-speaker English Libri\-Speech~\cite{panayotov2015librispeech} corpus of read audiobooks. Models are trained on forced alignments of the 960-hour training set and evaluated on dev-clean (5.4 hours from 40 speakers).
On this data, I compare the bitrates of phones, syllables, words, multi-word combinations obtained from BPE at the word-token level, as well as intermediate units. The number of tokens and types for phones, syllables and words in the training data are shown in Table~\ref{tbl:unit_stats}.
%I now provide details for the different units.

The phone-level forced alignments~\cite{mcauliffe2017montreal} use an ARPABet phoneset without stress markers, giving a set of 39 phones.
To obtain syllable units from the phone-level forced alignments, a rule-based method is used~\cite{gorman2013syllabify}, yielding within-word syllables, as was also done in~\cite{visser2026zerosyl}.

\begin{table}[!t]
    \caption{Number of tokens and types for each unit on the Libri\-Speech 960h training set.}
    \label{tbl:unit_stats}
    \mytable
    \begin{tabularx}{\linewidth}{@{}LRRr@{}}
        \toprule
        Units & Tokens & Types & Non-singleton types \\
        \midrule
        Phones    & 33,464,416 & 39     & 39     \\
        Syllables & 13,025,947 & 13,656 & 11,386 \\
        Words     & 9,403,427  & 84,163 & 56,255 \\
        \bottomrule
    \end{tabularx}
%    \vspace*{-3pt}
\end{table}

%As illustrated in Figure~\ref{fig:setup}, w
Word types are defined based on their unique phonetic realisations, e.g.\ \wrd{two} and \wrd{too} are both represented as the type \phn{t-uw}. This is needed so that the exact phone sequence is recoverable from a word tokenisation -- a one-to-one correspondence necessary for proper bitrate comparisons.
For the word transformer models, only types occurring more than once in the training data are used in the output vocabulary, giving an output layer of 56k types (last column and row in Table~\ref{tbl:unit_stats}). Singletons are  treated as unseen and are spelled out as described above.

Maybe the most efficient units are something larger than words, like phrases or common multi-word combinations? To test this, I use byte pair encoding (BPE)~\cite{gage1994new,sennrich2016neural}, but
rather than applying standard BPE where co-occurring characters are combined, I apply BPE at the word-token level to find new multi-word combinations.
New types like \phn{ah-v-dh-ah} (\wrd{of the}) are therefore added to the vocabulary. I again constrain the starting vocabulary to words occurring at least twice (56k types), and then add the new merged units. I consider 200, 1,000 and 5,000 merges. For the transformer models, this increases the output vocabulary by the merge count. Unseen units are again spelled out as phones. With 5,000 merges, the longest unit is \phn{aa-n-dh-iy-ah-dh-er-s-ay-d-ah-v-dh-ah} (\wrd{on the other side of~the}).

\section{Results}

\subsection{Main results}

\begin{figure*}[!t]
    \centering
    \includegraphics{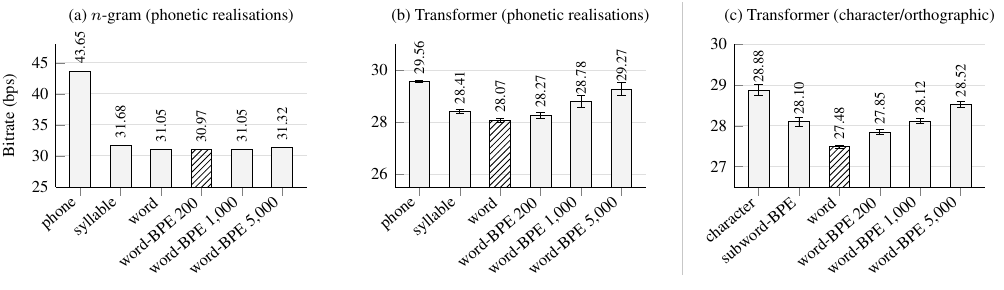}
    \vspace{-20pt}
    \caption{Bitrates for different linguistic units.
    In the main experiments, (a) $n$-gram and (b) transformer models are used to determine the bitrates of different tokenisations using phones as the atomic unit. In sanity-check experiments, (c) a transformer is used on different units using characters as the atomic unit.
%    : $n$-gram models on phone-based units (left), transformer models on phone-based units (middle) and transformer models on character-based units (right).
    The hatched bar is the best performance in each panel and error bars for the transformers are standard deviations over five seeds. Vertical axes differ between panels, as the bitrates are not directly comparable.}
    \label{fig:results}
\end{figure*}

The goal of this paper is to see which unit has the best information rate under a constrained modelling budget. For the main comparison, we look at phones, syllables, words and different word-BPE tokenisations (200, 1,000 and 5,000 merges).
We consider two model classes: third-order $n$-gram language models and transformer language models.
The results are respectively given in Figures~\ref{fig:results}a and \ref{fig:results}b.
Error bars for the transformer model are standard deviations across five training seeds.

Under both the $n$-gram and transformer models, we see that a unit larger than syllables but smaller than long multi-word combinations %(word-BPE 1,000 or 5,000)
gives the lowest bitrates. For the $n$-gram model, this is the word-BPE 200 units, while for the transformer, this is word units. The latter achieves the best overall bitrate of 28.07~bps among models trained on phonetic realisations.

Although the differences for the $n$-gram models are small, bootstrap confidence estimates~\cite{bisani2004bootstrap} reveal that all pairwise differences are significant at the 95\% level, except for the word vs the word-BPE 1,000 models (which makes sense given that their results are both 31.05 bps). Bootstrap confidence interval estimation also shows that pairwise differences between adjacent transformer models are statistically significant in all cases.\footnote{I use $B = \text{10,000}$ bootstrap samples at the speaker level, i.e.\ the 40 evaluation speakers are sampled with replacement with the corresponding utterances forming the bootstrap sample.} (This is despite some overlap in the error bars over different seeds, but these bars do not constitute a paired comparison.)

\subsection{Additional experiments and sanity checks}

My goal was to compare the information rates of the linguistic units above, but other (intermediate) units and comparisons are useful for supporting the above results and their interpretation.

The first of these uses the more standard form of BPE, where atomic units (normally characters but here I use phones) are merged to form new subword units. Concretely, I apply BPE to the phone units to obtain new types like \phn{ah-n} (\wrd{an}) and \phn{dh-ah} (\wrd{the}). I merge to get a vocabulary of 2,000 and disallow merges across word boundaries (as is standard in BPE). The hypothesis would be that the bitrate of this tokenisation would lie somewhere between phones and words (maybe close to syllables), and this is indeed the case: 31.59 bps under the $n$-gram and 28.98 $\pm$ 0.279 bps under the transformer. %Presumably, with this approach, we could get operating points anywhere between phones and words by adjusting the vocabulary accordingly.

The results so far are obtained using phones as the smallest atomic unit. As a sanity check, I repeat the analysis on characters. Words are now represented in their orthographic form so that \wrd{two} and \wrd{too} are now separate entries in the vocabulary. Results are given in Figure~\ref{fig:results}c when using transformers (20M parameters) to measure bitrate for characters, subword-BPE-merged characters (2,000 types), orthographic words, and word tokens merged with BPE (200, 1,000 and 5,000 merges). Bitrates are lower than those for phone-based models in Figure~\ref{fig:results}b, but these are not directly comparable: phones and characters do not have a direct one-to-one correspondence. Nevertheless, we see the same trend as with the phone models, with orthographic words giving significantly lower bitrates than the other units. % when based on characters.

Finally, I wanted to gauge how close we are to the true bitrate -- the lowest achievable. Given the discussion in Section~\ref{sec:units_dont_matter}, this clearly isn't possible since we will never have the true model. Nevertheless, we can consider bitrates when training data and model size are dramatically scaled.
%I cannot train models of this size myself, but I can use available open-weight models. 
%This would then tie the analysis to that model's units (so we cannot compare different tokenisations), but this is acceptable given that our aim now is not unit comparison but rather an upper bound estimate.
For this I use the 1.5B- and 3B-parameter Qwen2.5 models~\cite{qwen2024qwen2.5}. These models are trained on roughly 18T tokens of text. They use a byte-level BPE tokeniser with an output vocabulary of about 150k types. Because Qwen's training domain and punctuation are different from the normalised LibriSpeech text, I condition the Qwen models on a lowercase unpunctuated primer drawn from train-clean-100, prepended to every evaluation utterance. With this setup, Qwen2.5-1.5B achieves a bitrate of 25.90 bps, while Qwen2.5-3B achieves 25.26 bps. We can compare these to the character-based result in Figure~\ref{fig:results}c, where the best overall bitrate is 27.48 bps. This shows that substantially larger models can achieve even better bitrates. But the aim of this paper was a systematic comparison of different units \textit{under the same modelling budget}. %.\footnote{As model size and amount of data scales, we should expect the different units to give similar bitrates.}
In this light, the bitrates of the Qwen models serve as a reasonable sanity check on the bitrates achieved. % above.

\section{Discussion, limitations and future work}

The goal of this paper was to compare different linguistic units to see whether an optimal information rate exists when measured on the same data under the same constrained modelling budgets. Results when using phones as the atomic unit suggest that words (or potentially units just slightly longer than words) give a good trade-off, %between the number of types and unit size,
yielding better bitrates than syllables or very long multi-word combinations. These results are supported by experiments with characters as the atomic unit, where orthographic words (now spelled out as characters rather than phones) also achieve the lowest bitrates when using a 20M-parameter transformer language model.

The above seem to suggest that words are more information-theoretically efficient than smaller or much larger units. However, the concept of a ``word'' is controversial and not consistent across languages~\cite{haspelmath2011indeterminacy,haspelmath2023defining}.
This is a limitation of this study, and future work should  repeat a similar analysis on other languages. But my intention here is not to claim that ``words'' are the most efficient units. What I do claim is that the most efficient unit (under a limited budget) is something bigger than phones or syllables and smaller than long multi-word combinations. And I suspect that this will also hold on other languages.

In the introduction I noted that this work could have relevance to human language processing. A problem is that this paper measures the information rate of phonetic sequences rather than that of the message that a speaker intends to convey. Measuring the latter is far less straightforward. One possible route would be to adopt a psycholinguistic perspective, examining how different tokenisations relate to human language processing; \cite{oh2025impact} is one example in this direction. Ultimately, it remains an interesting open question whether and how an information-theoretic analysis like the one here can tell us something about the linguistic representations employed by humans.

\newpage
\section*{Acknowledgements}

I would like to thank Nicol Visser, Benjamin van Niekerk, Simon Malan, Camryn Abrahamson, Daanshil Ramgutty, Kyle Janse van Rensburg, and Danel Slabbert for helpful discussions.

\section*{AI disclosure}

The code for this project was written using AI (Claude Opus). I conceived of all the experiments and carefully laid out a plan; AI then implemented it. I reviewed the code, found and fixed bugs, edited the code to improve clarity, and wrote additional documentation. I wrote a first full draft of this paper before using any AI. I then used Claude, ChatGPT and Grammarly to find factual mistakes, fix spelling and grammar mistakes, highlight unclear sentences and propose alternatives. I then reviewed the entire paper again.

\bibliography{inforatebib}

\end{document}